\documentclass[letterpaper, 10 pt, conference]{ieeeconf}  % Comment this line out if you need a4paper

\IEEEoverridecommandlockouts                              % This command is only needed if 
\usepackage{hyperref}       % hyperlinks
\usepackage{url}            % simple URL typesetting
\usepackage{booktabs}       % professional-quality tables
\usepackage{amsfonts}       % blackboard math symbols
\usepackage{nicefrac}       % compact symbols for 1/2, etc.
\usepackage{microtype}      % microtypography
\usepackage{xcolor}         % colors

\usepackage{graphicx}
\usepackage{amsmath}
\usepackage{multirow}
\usepackage{multicol}
\usepackage{bbding}
\usepackage{wrapfig}

\usepackage{caption}
\usepackage{subcaption}
\usepackage{amssymb}
\usepackage{footnote}
\usepackage{url}
\usepackage{algorithm}
\usepackage{listings}
\usepackage{tabularx}
\usepackage {pifont} 
\usepackage{bm}
\usepackage{tablefootnote}
\usepackage{tikz}
\usetikzlibrary{matrix, backgrounds,fit}
\usepackage{lipsum}
\usepackage{algpseudocode}
\usepackage{makecell}
\usepackage{amssymb}
\usepackage{multirow}
\usepackage{colortbl}
\usepackage{xspace}
\def\ourmethod{UniFuture\xspace}

\hypersetup{
    colorlinks=true,
    linkcolor=red,
    filecolor=magenta,      
    urlcolor=magenta,
    citecolor=kaiming-green
}

\title{\LARGE \bf
UniFuture: A 4D Driving World Model for Future Generation and Perception
}

\author{Dingkang Liang$^{1}$, Dingyuan Zhang$^{1}$, Xin Zhou$^{1}$, Sifan Tu$^{1}$, \\ Tianrui Feng$^{1}$, Xiaofan Li$^{2}$, Yumeng Zhang$^{2}$, Mingyang Du$^{1}$, Xiao Tan$^{2}$, Xiang Bai$^{\dag1}$
\thanks{$^{1}$ Huazhong University of Science and Technology.} % 请根据实际情况确认学院英文名称
\thanks{$^{2}$Baidu Inc., China.} % 请确认单位2的信息
\thanks{\dag Corresponding author. {\tt\small dkliang@hust.edu.cn}}
}
\begin{document}

\maketitle
\thispagestyle{empty}
\pagestyle{empty}

%%%%%%%%%%%%%%%%%%%%%%%%%%%%%%%%%%%%%%%%%%%%%%%%%%%%%%%%%%%%%%%%%%%%%%%%%%%%%%%%
\begin{abstract}
We present \ourmethod, a unified 4D Driving World Model designed to simulate the dynamic evolution of the 3D physical world. Unlike existing driving world models that focus solely on 2D pixel-level video generation (lacking geometry) or static perception (lacking temporal dynamics), our approach bridges appearance and geometry to construct a holistic 4D representation. Specifically, we treat future RGB images and depth maps as coupled projections of the same 4D reality and model them jointly within a single framework. To achieve this, we introduce a Dual-Latent Sharing (DLS) scheme, which maps visual and geometric modalities into a shared spatio-temporal latent space, implicitly entangling texture with structure. Furthermore, we propose a Multi-scale Latent Interaction (MLI) mechanism, which enforces bidirectional consistency: geometry constrains visual synthesis to prevent structural hallucinations, while visual semantics refine geometric estimation. During inference, \ourmethod~can forecast high-fidelity, geometrically consistent 4D scene sequences (image-depth pairs) from a single current frame. Extensive experiments on the nuScenes and Waymo datasets demonstrate that our method outperforms specialized models in both future generation and geometry perception, highlighting the efficacy of unified 4D modeling for autonomous driving. The code is available at \url{https://github.com/dk-liang/UniFuture}.
\end{abstract}

%%%%%%%%%%%%%%%%%%%%%%%%%%%%%%%%%%%%%%%%%%%%%%%%%%%%%%%%%%%%%%%%%%%%%%%%%%%%%%%%
\section{INTRODUCTION}
\label{sec:intro}

The physical world in which autonomous vehicles operate is inherently four-dimensional, consisting of 3D spatial geometry evolving over the temporal dimension. Consequently, an ideal Driving World Model (DWM) should be capable of simulating this 4D dynamic evolution, enabling the ego-vehicle to anticipate how the 3D surroundings will unfold. While traditional DWMs~\cite{wang2024drivedreamer,gao2024vista,zhou2025hermes,tu2025role} have made significant strides in simulating future scenarios, they often fall short of capturing the full 4D nature of the environment.

\begin{figure}[!t]
    \centering
    \includegraphics[width=0.475\textwidth]{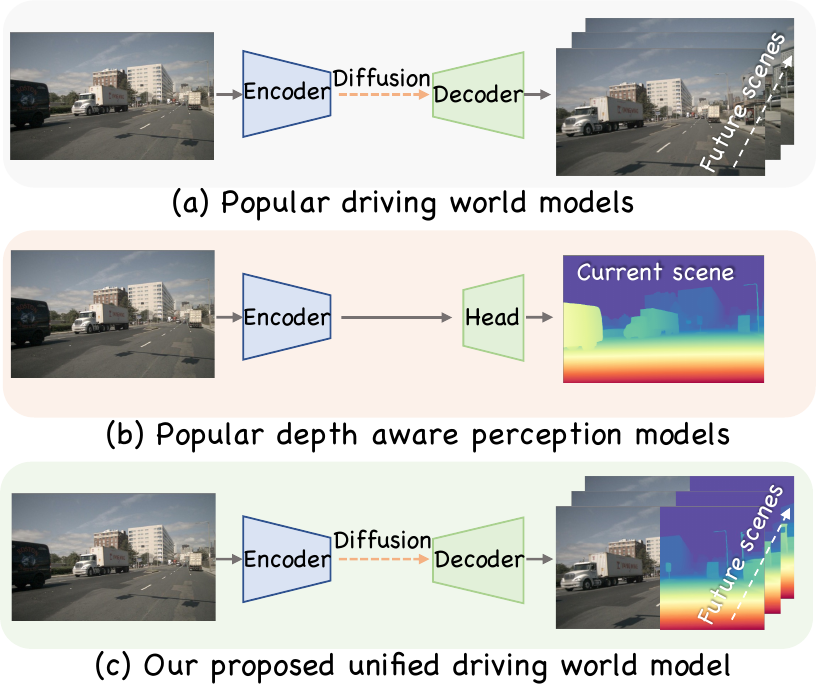}
    \caption{An intuition comparison between existing methods and our unified world model. Unlike (a) 2D-focused generation and (b) static 3D perception, our \ourmethod (c) jointly predicts future appearance and geometry, effectively modeling the 4D evolution of the world.}
    \label{fig:intro}

\end{figure}

Recent advancements in DWMs, driven by large-scale pre-trained diffusion models, have predominantly focused on fine-grained 2D video generation~\cite{gao2024vista,li2024drivingdiffusion,yang2024genad,wang2024driving,wen2024panacea}. As shown in Fig.~\ref{fig:intro}(a), these models excel at synthesizing visually realistic RGB sequences. However, they largely overlook the underlying 3D geometry, such as depth, essentially predicting cinematic hallucinations rather than physical realities. Without explicit geometric modeling, these 2D-based DWMs struggle with spatial reasoning tasks, including handling occlusions and estimating accurate distances. This results in predictions that may be visually plausible but physically inconsistent. Conversely, depth-aware perception models~\cite{piccinelli2024unidepth,yang2024depth,duan2024diffusiondepth} excel at extracting high-level geometric structures but are typically limited to static 3D snapshots of the present or past. As illustrated in Fig.~\ref{fig:intro}(b), they lack the capability to forecast how these 3D structures will evolve, missing the critical temporal dimension. These limitations point to a significant gap: \textit{Can we develop a unified driving world model that integrates appearance, geometry, and dynamics to forecast the authentic 4D evolution of driving scenes?}

To bridge this gap, we propose \ourmethod, a unified 4D Driving World Model that simultaneously forecasts future scene appearance (RGB) and geometry (Depth), as shown in Fig.~\ref{fig:intro}(c). We posit that an image and its corresponding depth map are merely distinct projections of the same underlying 4D reality. Therefore, modeling them in isolation is suboptimal. To unify them, we introduce a Dual-Latent Sharing (DLS) scheme. Rather than training separate encoders for texture and geometry, DLS maps both modalities into a shared latent space. Implicitly, this constructs a unified spatio-temporal representation. This design not only eliminates the need for additional pre-training but also ensures that the generated geometry and texture are entangled at the feature level, mirroring their physical correlation in the real world.

Building on this unified representation, we propose a Multi-scale Latent Interaction (MLI) mechanism to enforce spatio-temporal consistency. In a 4D world, geometry constrains visual appearance, while visual cues refine geometric estimation. MLI facilitates this bidirectional information flow within a multi-scale UNet architecture. By iteratively injecting depth latents into the image stream and propagating refined image features back to the depth stream, we ensure that the predicted future is not just a video, but a coherent 4D point cloud sequence (as demonstrated in our experiments). This mutual interaction integrates low-level pixel synthesis with high-level spatial reasoning, resulting in predictions that are geometrically consistent over time.

Our approach shifts from 2D video prediction to 4D world modeling, offering several key advantages. First, the synergy of appearance and geometry enhances prediction quality. Specifically, depth priors stabilize video generation against temporal distortions, while appearance priors refine the details of future depth estimation. Second, our unified framework serves as a powerful foundation for downstream tasks. It enables self-driving systems to simulate different scenarios and make informed decisions~\cite{li2025driverse}, while also generating highly consistent annotated data for automated training. Extensive experiments on the nuScenes dataset demonstrate that \ourmethod achieves state-of-the-art performance. Compared to the strong baseline Vista~\cite{gao2024vista}, we reduce FID by 23.9\% while simultaneously outperforming specialized diffusion-based depth estimation methods~\cite{ke2024repurposing} in future geometric prediction.

The main contributions are summarized as follows: \textbf{1)} We propose \ourmethod, a novel 4D Driving World Model framework. By seamlessly integrating future generation and perception, we extend world modeling from 2D pixel space to 4D geometric space. \textbf{2)} We introduce the Dual-Latent Sharing (DLS) scheme and Multi-scale Latent Interaction (MLI) mechanism. These modules effectively unify heterogeneous modalities in a shared latent space and enforce bidirectional spatio-temporal consistency. \textbf{3)} Our method achieves impressive performance in both future scene generation and depth estimation, demonstrating the potential of unified 4D modeling for autonomous driving.

\section{RELATED WORK}
\subsection{World Models in Autonomous Driving} 
World models predict scene evolution based on current observations~\cite{ha2018world, guan2024world_survey} and have become a key focus in autonomous driving. Many approaches~\cite{kim2021drivegan, lu2024wovogen, li2024drivingdiffusion} excel in forecasting 2D visual representations, demonstrating strong dynamic modeling capabilities. Most methods rely on generative techniques like autoregressive transformers~\cite{hu2023gaia, chen2024drivinggpt} and diffusion models~\cite{guo2024infinitydrive, jiang2024dive, wen2024panacea_plus} for future video prediction. Specifically, GAIA-1~\cite{hu2023gaia} formulates world modeling as a sequence task using an autoregressive transformer, while ADriver-I~\cite{jia2023adriveri} integrates multi-modal LLMs and diffusion for control and frame generation. DriveDreamer~\cite{wang2024drivedreamer} enforces structured traffic constraints in diffusion-based prediction, with DriveDreamer-2~\cite{zhao2024drivedreamer} further incorporating LLMs for customizable video generation. Drive-WM~\cite{wang2024drive_wm} enhances multi-view consistency via view factorization, and GenAD~\cite{yang2024genad} introduces large-scale video datasets to improve zero-shot generalization. HERMES~\cite{zhou2025hermes} seamlessly integrates 3D scene understanding and future scene evolution (generation) in an MLLM. 
Based on GenAD, Vista~\cite{gao2024vista} enhances dynamics and preserves structural details with two extra loss functions, achieving high-resolution, high-fidelity, and long-term scene evolution prediction. Subsequent works~\cite{hassan2024gem,guo2024infinitydrive} have made significant advancements in prediction duration~\cite{guo2024infinitydrive} and the integration of prediction with planning~\cite{chen2024drivinggpt,wang2024driving,liang2025cook}. Despite advancements, these world models only focus on low-level visual representations and overlook the geometry information, which hinders the ability of spatial reasoning. Instead, the proposed UniFuture integrates visual information with geometry features seamlessly, enabling the geometry-aware reasoning ability of the world model.

\subsection{Monocular Depth Estimation}
Monocular depth estimation, which aims to recover the geometry information from a single image or video, has gained significant attention in autonomous driving research~\cite{rajapaksha2024deep}. It serves as a fundamental step for various downstream 3D tasks~\cite{li2023dds3d,liang2024pointmamba,zhou2024dynamic,liang2025parameter,zhang2023simple}. Conventional works~\cite{song2021monocular,bhat2021adabins,yuan2022neural} leverage multi-scale feature fusion to combine global and local depth information, achieving remarkable performance within specific domains. MiDaS~\cite{ranftl2020towards} and ZoeDepth~\cite{bhat2023zoedepth} leverage multi-dataset joint training to enhance generalization, while DepthAnything~\cite{yang2024depth,yang2024depth2} explores large-scale unlabeled and synthetic data to improve detail modeling in complex scenes.
Another promising direction tries to incorporate rich priors of generative models trained on vast amounts of wild images. Marigold~\cite{ke2024repurposing} pioneers the use of Stable Diffusion~\cite{rombach2022high} for affine-invariant depth estimation, followed by Lotus~\cite{he2024lotus}, which refines denoising schedules for better adaptation. DepthCrafter~\cite{hu2024depthcrafter} extends this framework to open-world videos for consistent long-sequence depth estimation. MERGE~\cite{lin2025more} proposes a unified model for generation and depth estimation, starting from
a fixed pre-trained text-to-image model. While existing depth estimation methods focus on perceiving the current or past environment, our approach integrates future depth estimation with the world model. This synergy leverages the world model’s reasoning capabilities for high-quality depth prediction while using precise geometric cues to enhance dynamic scene modeling.

\section{METHOD}

\begin{figure*}[t]
	\begin{center}
		\includegraphics[width=0.97\linewidth]{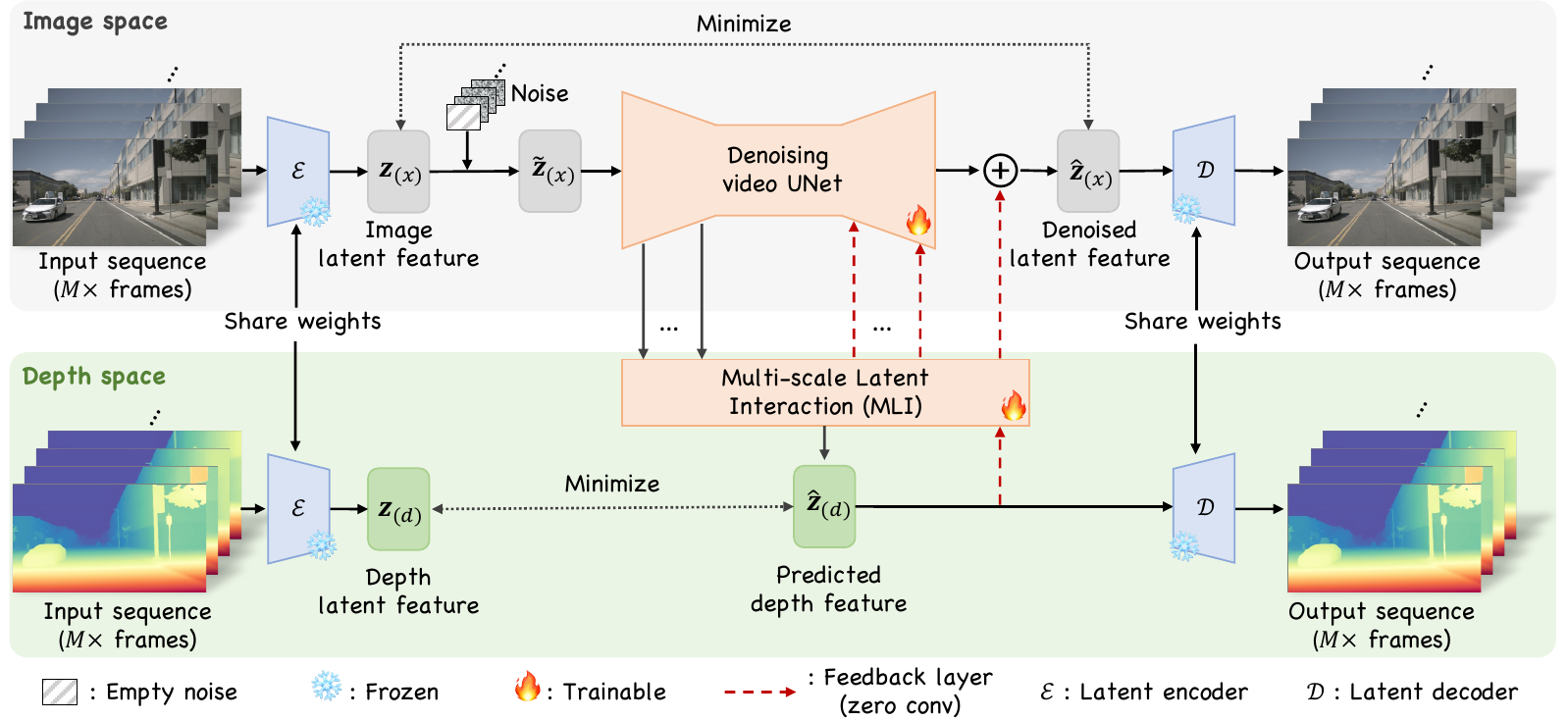}
	\end{center}
         \vspace{-5pt}
    \caption{The training pipeline of \ourmethod. To learn a 4D Driving World Model, we introduce the Dual-Latent Sharing (DLS) scheme, which unifies visual appearance (image) and 3D geometry (depth) into a shared latent space without additional pre-training. The image latent undergoes a conditional denoising process, while the depth latent is explicitly predicted via the Multi-scale Latent Interaction (MLI) mechanism. MLI enforces bidirectional spatio-temporal consistency between texture and structure, ensuring coherent 4D scene evolution.}
	\label{fig:pipeline}

\end{figure*}

\begin{figure}[t]
    \centering
    \includegraphics[width=0.97\linewidth]{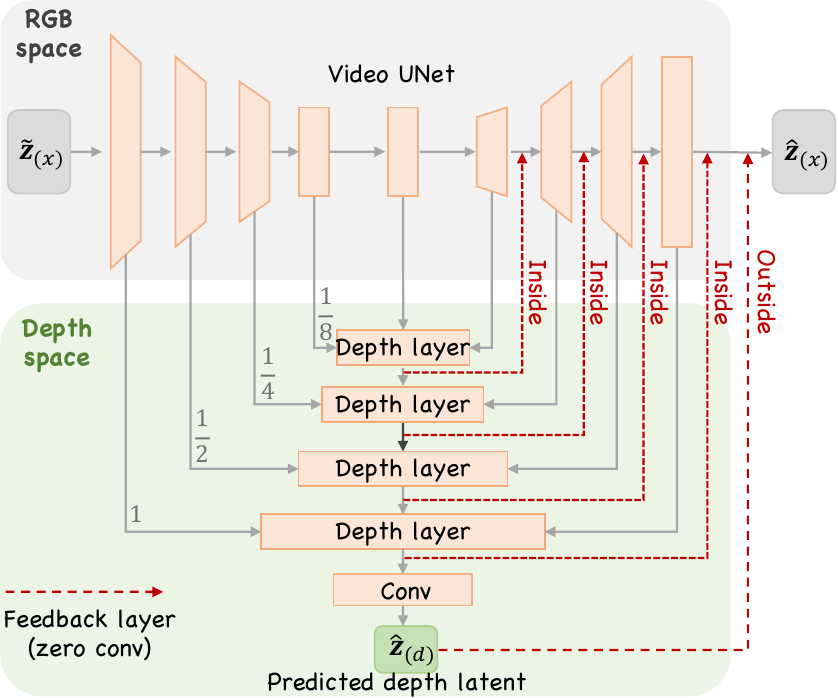}
    \caption{The details of our proposed Multi-scale Latent Interaction (MLI) mechanism, designed to bridge the gap between visual and geometric feature spaces.}
    \label{fig:multi_scale}

\end{figure}

This paper proposes \ourmethod, a unified 4D Driving world model that seamlessly integrates future scene generation with depth-aware perception to simulate the dynamic evolution of driving environments. Built upon an SVD-based video generation framework~\cite{gao2024vista}, our approach bridges the gap between appearance (RGB) and geometry (Depth), which are two heterogeneous projections of the same 4D reality. We achieve this through two key components: 1) A \textbf{Dual-Latent Sharing (DLS)} scheme, which maps both modalities into a unified spatio-temporal latent space, implicitly entangling texture with geometry; and 2) A \textbf{Multi-scale Latent Interaction (MLI)} mechanism, a bidirectional feedback system that refines features across scales. These components ensure that pixel-level synthesis is geometrically grounded, and spatial reasoning is visually informed, resulting in coherent 4D predictions.

\begin{figure}[t]
	\begin{center}
		\includegraphics[width=0.97\linewidth]{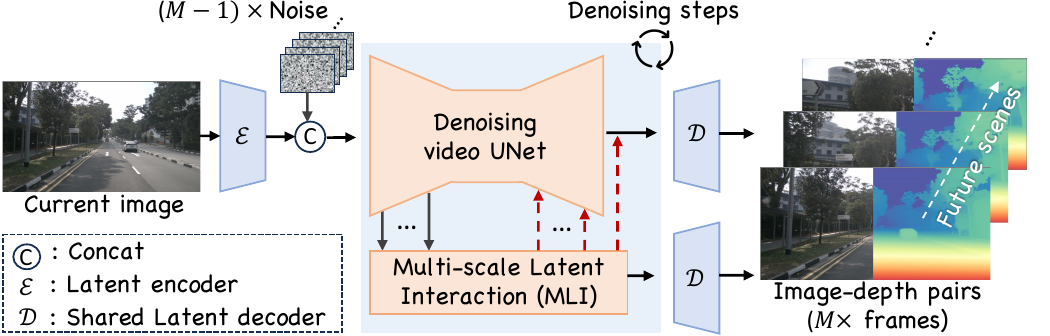}
	\end{center}
         \vspace{-5pt}
	\caption{The inference pipeline of \ourmethod. It transforms a single 2D observation into a 4D forecast (future image-depth pairs). The latent representation evolves through the MLI-enhanced denoising UNet, producing geometrically aligned future sequences.}
	\label{fig:inference}
   
\end{figure}

During training (Fig.~\ref{fig:pipeline}), \ourmethod takes an image-depth pair sequence with $M$ frames as input, treating it as a discretized 4D scene representation. The image latent feature follows the conditional denoising process\footnote{The first frame serves as a noise-free condition.}, while the depth latent feature is concurrently optimized. During inference (Fig.~\ref{fig:inference}), given only a single current image, the model predicts future image-depth pairs by concatenating $(M-1) \times$ noise embeddings. This process effectively hallucinates a temporally and geometrically consistent 4D future from a static 3D observation.

\subsection{Dual-Latent Sharing Scheme: Constructing a Unified 4D Latent Space}

In the physical world, an RGB image and its corresponding depth map are intrinsically linked descriptions of the same scene. We argue that learning a 4D world model requires a representation where these two modalities are aligned. To this end, we introduce the Dual-Latent Sharing (DLS) scheme. Rather than treating depth estimation as an auxiliary task with a separate encoder, DLS unifies image and depth representations within a shared latent space.

As shown in Fig.~\ref{fig:pipeline}, given an image sequence $\boldsymbol{x}$ (appearance) and its corresponding depth maps $\boldsymbol{d}$ (geometry), we process both through a shared pre-trained latent encoder $\mathcal{E}$, obtaining $\boldsymbol{z}_{(x)}$ and $\boldsymbol{z}_{(d)}$. The image latent $\boldsymbol{z}_{(x)}$ undergoes the standard diffusion process to model temporal dynamics. Concurrently, the depth latent $\hat{\boldsymbol{z}}_{(d)}$ is derived via the Multi-scale Latent Interaction (MLI) mechanism. Finally, both are reconstructed back to the pixel/metric space using the shared decoder $\mathcal{D}$. 

By forcing depth maps to traverse the same latent space as natural images, we implicitly encode geometry using the rich, pre-trained semantic priors of the video generator. This eliminates the need for additional depth-specific pre-training and enables seamless cross-modal feature flow.

\subsection{Multi-scale Latent Interaction: Enforcing Spatio-Temporal Consistency}
To ensure that the generated 4D predictions are physically consistent (i.e., textures adhere to surfaces, and shapes do not deform unrealistically over time), we propose the Multi-scale Latent Interaction (MLI) mechanism. MLI facilitates explicit bidirectional information flow between the appearance stream and the geometry stream. As illustrated in Fig.~\ref{fig:multi_scale}, MLI consists of hierarchical depth layers for alignment, followed by ``Inside'' and ``Outside'' feedback loops.

\textbf{Depth Layers for Feature Alignment.} To translate the rich semantic features from the video UNet into geometric representations, we introduce a hierarchical fusion strategy. Specifically, we extract multi-scale features from the UNet encoder and decoder at scales \{$1$, $\frac{1}{2}$, $\frac{1}{4}$, $\frac{1}{8}$\}. Each depth layer aggregates these features with the upsampled output from the preceding layer. This progressive fusion ensures that the depth estimation benefits from both high-level semantic context (essential for understanding object extent) and low-level structural details (essential for boundaries).

\textbf{Inside Feedback: Geometry-Guided Generation.} While image and depth share a latent space, their feature distributions differ. To bridge this, we apply a zero-initialized convolution to the intermediate depth latent feature $\mathbf{X}$ before injecting it back into the video generation stream.
As indicated by the red lines in Fig.~\ref{fig:multi_scale}, inside feedback flows from the intermediate depth layers to the corresponding UNet stages:
\begin{equation}
\label{eq:zero}
\mathbf{Y}=\mathtt{ZeroConv}\left(\mathbf{X}\right),
\end{equation} 
where $\mathbf{Y}$ acts as a geometric condition added to the video UNet features. By initializing weights to zero, the model starts with standard video generation and progressively learns to utilize geometric cues to refine texture synthesis, preventing geometric inconsistencies in the generated video.

\textbf{Outside Feedback: Texture-Refined Geometry.} To further align the final output, we introduce an outside feedback mechanism. The final predicted depth latent $\hat{\boldsymbol{z}}_{(d)}$ is injected into the denoised image latent $\hat{\boldsymbol{z}}_{(x)}$. This step ensures that the final appearance is strictly conditioned on the predicted geometry, reinforcing the structural integrity of the generated 4D scene.

By integrating these components, MLI achieves a closed-loop interaction: geometry constrains appearance, and appearance refines geometry. This allows \ourmethod~ to simultaneously generate high-fidelity image-depth pairs that are temporally coherent and geometrically accurate.

\subsection{Training Objectives}

Our training objectives enforce consistency in both the latent and pixel spaces, crucial for 4D modeling. At the latent level, we minimize the reconstruction error of the denoised image latent $\hat{\boldsymbol{z}}_{(x)}$ using a composite loss $\mathcal{L}_{(x)}$ (including MSE and structural losses), following Vista~\cite{gao2024vista}. Similarly, a loss $\mathcal{L}_{(d)}$ constrains the depth latent. To ensure the physical validity of the predicted geometry, we further impose a Scale- and Shift-Invariant loss $\mathcal{L}_{SSI}$ between the reconstructed predicted depth and the ground truth, following Depth Anything~\cite{yang2024depth}. The overall objective is:
\begin{equation}
    \mathcal{L} =\mathcal{L}_{(x)} + \mathcal{L}_{(d)} + \lambda\cdot \mathcal{L}_{SSI},
\end{equation}
where $\lambda$ balances the contributions.

\subsection{Inference Phase}

Unlike previous approaches that treat prediction and perception as separate tasks, \ourmethod leverages a single current frame to forecast a consistent 4D future. As illustrated in Fig.~\ref{fig:inference}, the input image is encoded and concatenated with $(M-1) \times$ Gaussian noise maps. This combined volume is then refined through the MLI-enhanced UNet. The process jointly evolves the appearance and geometry latents over time, which are finally decoded into a sequence of image-depth pairs. This resulting sequence constitutes a comprehensive 4D world representation, maintaining perceptual realism and structural coherence crucial for downstream autonomous driving applications.

\section{EXPERIMENTS}

\begin{table*}[t]
    \footnotesize
    \centering
    \caption{Comparison of our method with specialized generation and depth estimation models. $^{\bigstar}$ marks our baseline method, which is fine-tuned under the same resolution and iteration constraints using the official pre-trained weight of Vista~\cite{gao2024vista}. 
    Our method predicts depth up to 25 future frames, while Marigold~\cite{ke2024repurposing} supports only single-frame depth estimation. $^{\blacklozenge}$  To enable a reasonable comparison, we train Marigold with next-frame supervision using the 1st and 12th future frames. }
    \vspace{-7pt}
    \setlength{\tabcolsep}{4mm}
    
    \begin{tabular}{lcccccccc}
    \toprule
     \multirow{2.3}*{Method} & \multirow{2.3}*{Reference} & \multirow{2.3}*{Resolution} & \multicolumn{2}{c}{Generation} & \multicolumn{4}{c}{Depth estimation}\\
     \cmidrule(lr){4-5}
     \cmidrule(l){6-9}
     & & & FID $\downarrow$ & FVD $\downarrow$ & AbsRel $\downarrow$ & $\delta_1$ $\uparrow$ & $\delta_2$ $\uparrow$ & $\delta_3$ $\uparrow$\\
     \midrule
    \rowcolor{gray!20}  \multicolumn{9}{c}{Only future depth estimation}\\
     \midrule 

    Marigold (0-th frame)~\cite{ke2024repurposing} & CVPR 24 & 320$\times$576 &  \multicolumn{2}{c}{\multirow{3}{*}{Unsupported}} & 20.4 & 80.3 & 93.1 & 96.5 \\
     Marigold (1-st frame)$^{\blacklozenge}$~\cite{ke2024repurposing} & CVPR 24 & 320$\times$576 & &  & 21.9 & 77.7 & 91.2 & 95.6   \\
    Marigold (12-th frame)$^{\blacklozenge}$~\cite{ke2024repurposing} & CVPR 24 & 320$\times$576 & & & 39.0 & 65.8 & 81.9 & 89.7   \\

     \midrule
    \rowcolor{gray!20} \multicolumn{9}{c}{Only future generation}\\
     \midrule
    DriveGAN~\cite{kim2021drivegan} & CVPR 21 & 256$\times$256 & 73.4 & 502.3 & \multicolumn{4}{c}{\multirow{8}{*}{Unsupported}} \\
    DriveDreamer~\cite{wang2024drivedreamer} & ECCV 24 & 128$\times$192 & 52.6 & 452.0 \\
    WoVoGen~\cite{lu2024wovogen} & ECCV 24 & 256$\times$448 & 27.6 & 417.7 \\
    DrivingDiffusion~\cite{li2024drivingdiffusion} & ECCV 24 & 512$\times$512 & 15.8 & 332.0 \\
    GenAD~\cite{yang2024genad} & CVPR 24 & 256$\times$448 & 15.4 & 184.0 \\
    Panacea~\cite{wen2024panacea} & CVPR 24 & 256$\times$512 & 17.0 & 139.0 \\
    Drive-WM~\cite{wang2024driving} & CVPR 24 & 192$\times$384 & 15.2 & 122.7 \\

    Vista$^{\bigstar}$~\cite{gao2024vista} (baseline) & NeurIPS 24& 320$\times$576 & 15.5 & 101.5\\
    \midrule
   \rowcolor{gray!20}  \multicolumn{9}{c}{\textbf{Unify future generation and depth estimation}}\\
     \midrule
     \ourmethod~(ours) &-& 320$\times$576 & \textbf{11.8} & \textbf{99.9} & \textbf{8.936} & \textbf{91.4} & \textbf{97.6} &\textbf{98.9}\\
\bottomrule 
    \end{tabular}
    \label{tab:main_results}
    \vspace{-10pt}
\end{table*}
\subsection{Dataset and Evaluation Metric}

\textbf{Datasets.} We conduct experiments on the nuScenes dataset~\cite{caesar2020nuscenes}, a large-scale autonomous driving benchmark that provides multi-modal sensor data, including RGB images, LiDAR point clouds, and scene annotations. It contains 1,000 driving sequences collected in diverse urban environments. Following Vista~\cite{gao2024vista}, we use the front-camera RGB frames as the main training data. We also evaluate our model on the Waymo~\cite{sun2020scalability} dataset under a zero-shot setting to assess its generalization to unseen driving scenes. Since neither nuScenes~\cite{caesar2020nuscenes} nor Waymo~\cite{sun2020scalability} provides dense pixel-wise depth annotations, we adopt DepthAnythingV2~\cite{yang2024depth2} to generate depth labels.

\textbf{Evaluation metrics.} For the generation task, we evaluate the quality and realism of generated frames using Fréchet Inception Distance (FID) and Fréchet Video Distance (FVD), which measure the distributional similarity between generated and real images/videos. Lower FID and FVD scores indicate better generation quality. For the depth estimation task, we use Absolute Relative Error (AbsRel) to quantify relative depth differences and threshold accuracy ($\delta$) to measure the proportion of accurate predictions within a specified relative error. Higher $\delta$ values and lower AbsRel values indicate better depth estimation performance. It should be noted that to evaluate the consistency between the predicted future depth and the future scene, we utilize DepthAnythingV2~\cite{yang2024depth2} to process the predicted future scene as pseudo-depth labels when computing the aforementioned depth metrics unless otherwise specified.

\subsection{Implementation Details}
We adopt the representative framework of Vista~\cite{gao2024vista} as the baseline. Our model is trained with a batch size of 1 on 8 $\times$ NVIDIA H20 GPUs. We use the AdamW optimizer with a learning rate of $5 \times 10^{-5}$, and the training process runs for 8K iterations to ensure convergence. Besides, following Vista~\cite{gao2024vista}, we use the Exponential Moving Average (EMA) strategy to make training stable and drop out each activated action mode with a ratio of 15\% to allow classifier-free guidance. To optimize memory usage, we adopt the DeepSpeed ZeRO-2 strategy, which effectively reduces memory overhead. We set the length of the video $M=25$, and the loss balance weight $\lambda = 0.5$.

\subsection{Main Results}

We evaluate \ourmethod~on both future scene generation and future geometry perception to investigate its capability as a unified 4D world model. Unlike prior works that treat visual synthesis and depth estimation as disjoint objectives, \ourmethod~aims to simulate the holistic 4D evolution of driving scenes through shared representations and structured interaction. For evaluation, we adopt Vista~\cite{gao2024vista}, a state-of-the-art driving world model, as our primary baseline. As shown in Tab.~\ref{tab:main_results}, our method delivers consistently superior performance, validating our 4D modeling hypothesis.

\textbf{On future scene generation,} \ourmethod~achieves a significant 3.7-point reduction in FID (from 15.5 to 11.8) and a highly competitive FVD score compared to Vista~\cite{gao2024vista}. These improvements highlight the critical role of geometry-aware synthesis: by explicitly modeling depth in the shared latent space, our model enforces structural constraints on the generated video. This prevents common artifacts like object deformation or temporal flickering seen in pure 2D models~\cite{gao2024vista,li2024drivingdiffusion,wen2024panacea}, which rely solely on pixel-level patterns. In essence, \ourmethod~does not just ``paint'' pixels; it ``renders'' a physically grounded 4D world, leading to superior realism and temporal coherence.

\begin{table}[t]
    \centering
    \footnotesize
    \caption{Zero-shot results on the Waymo dataset. Our method demonstrates superior generalization in both 4D generation and perception.}
        \vspace{-7pt}
    \setlength{\tabcolsep}{1.2mm}
    \begin{tabular}{ccccccc}
    \toprule %[2pt] 
    \multirow{2.3}*{Method} & \multicolumn{2}{c}{Generation} & \multicolumn{4}{c}{Perception}\\
    \cmidrule(lr){2-3} 
    \cmidrule(lr){4-7}
     & FID $\downarrow$ & FVD $\downarrow$ & AbsRel $\downarrow$ & $\delta_1$ $\uparrow$ & $\delta_2$ $\uparrow$ & $\delta_3$ $\uparrow$\\
    \midrule
    % Vista & 14.0 & 198.5 & - & - & - & - \\
    Vista~\cite{gao2024vista} (baseline) & 23.8 & 238.4 &  \multicolumn{4}{c}{Unsupported} \\ 
    \ourmethod~(ours) & \textbf{16.3} & \textbf{227.6}  & \textbf{9.517} & \textbf{89.3} & \textbf{97.0} & \textbf{98.7 }\\
\bottomrule %[2pt]    
    \end{tabular}
    \label{tab:zero_shot}
    \vspace{-5pt}
\end{table}

\begin{table*}[!t]
  \centering
  \scriptsize
  \renewcommand\arraystretch{1.1}
  \caption{Comprehensive ablation studies validating the core components of our 4D world model: 
    (a) optimization paradigms, (b) unified latent representation (DLS), (c) multi-scale interactions, 
    (d) bidirectional feedback mechanisms, and (e) feature alignment strategies.}
  \label{tab:all_ablation}

  % ==== 第一排： (a) 和 (b) ====
  \begin{subtable}[t]{0.49\linewidth}
    \centering
    \caption{Optimization paradigms}
    \vspace{-5pt}
    \setlength{\tabcolsep}{1.5mm}
\begin{tabular}{@{}lrrcccc@{}}
    \toprule
     \multirow{2.3}*{Setting}          & \multicolumn{2}{c}{Generation} & \multicolumn{4}{c}{Depth estimation} \\
    \cmidrule(lr){2-3} \cmidrule(lr){4-7}
                    & FID $\downarrow$ & FVD $\downarrow$ 
                    & AbsRel $\downarrow$ & $\delta_1$ $\uparrow$ 
                    & $\delta_2$ $\uparrow$ & $\delta_3$ $\uparrow$ \\
    \midrule
    Image-only      & 15.5              & 101.5             
                    & \multicolumn{4}{c}{Unsupported}                             \\
    Depth-only      & 271.0             & 3143.2            
                    & 24.196              & 62.0                & 84.5                & 92.4               \\
    Detach-grad     & \textbf{11.2}     & 104.9             
                    & 12.351              & 85.2                & 95.2                & 97.8               \\
    Joint training & 11.8              & \textbf{99.9}    
                    & \textbf{8.936}      & \textbf{91.4}       & \textbf{97.6}       & \textbf{98.9}    \\
    \bottomrule
    \label{tab:optimization}
  \end{tabular}
  \end{subtable}
  \hfill
  \begin{subtable}[t]{0.48\linewidth}
    \centering
    \caption{Depth decoders (Validating DLS)}
    \vspace{-5pt}
    \setlength{\tabcolsep}{1.3mm}
    \renewcommand\arraystretch{2.0}
    \begin{tabular}{@{}lrrrcccc@{}}
      \toprule
      Decoder              & FID $\downarrow$ & FVD $\downarrow$ 
                           & AbsRel $\downarrow$ & $\delta_1$ $\uparrow$ 
                           & $\delta_2$ $\uparrow$ & $\delta_3$ $\uparrow$ \\
      \midrule
      Convention & 15.6  & 121.1    & 11.874            & 86.4    & 95.9    & 98.1 \\
      Our DLS              & \textbf{11.8} & \textbf{99.9} 
                           & \textbf{8.936}    & \textbf{91.4} & \textbf{97.6} & \textbf{98.9} \\
      \bottomrule
      \label{tab:vpd_like}
    \end{tabular}
  \end{subtable}

  % \vspace{2mm} % <- 强制第一排与第二排之间留空

  \setlength\tabcolsep{2pt}
  % ==== 第二排： (c), (d), (e) ====
  \begin{subtable}[t]{0.32\linewidth}
    \centering
    \caption{Multi-scale interactions}
        \vspace{-5pt}
    \renewcommand\arraystretch{1.36}
    \begin{tabular}{@{}lcccc@{}}
      \toprule
      Scale                             & FID $\downarrow$ & FVD $\downarrow$ 
                                        & AbsRel $\downarrow$ & $\delta_1$ $\uparrow$ \\
      \midrule
      1                                 & 12.9 & 114.4 & 9.677  & 91.1 \\
      1, $\tfrac12$, $\tfrac14$         & 12.6 & 106.9 & 9.277  & \textbf{91.4} \\
      1, $\tfrac12$, $\tfrac14$, $\tfrac18$ 
                                        & \textbf{11.8} & \textbf{99.9} 
                                        & \textbf{8.936} & \textbf{91.4} \\
      \bottomrule
      \label{tab:multiscale}
    \end{tabular}
  \end{subtable}
  \hfill
  \begin{subtable}[t]{0.32\linewidth}
    \centering
    \caption{Inside vs.\ outside feedback}
    \vspace{-5pt}
    \begin{tabular}{@{}cccccc@{}}
      \toprule
      In   & Out  & FID $\downarrow$ & FVD $\downarrow$ 
           & AbsRel $\downarrow$ & $\delta_1$ $\uparrow$ \\
      \midrule
      –    & –    & 13.3  & 123.3 & 9.034  & 91.5 \\
      \checkmark & –    & 12.2  & 110.1 & 9.034  & \textbf{91.6} \\
      –    & \checkmark & 13.5 & 119.0 & 9.100 & 91.4 \\
      \checkmark & \checkmark & \textbf{11.8} & \textbf{99.9} 
                 & \textbf{8.936} & 91.4 \\
      \bottomrule
      \label{tab:feedback}
    \end{tabular}
  \end{subtable}
  \hfill
  \begin{subtable}[t]{0.33\linewidth}
    \centering
    \caption{Feedback layer types}
        \vspace{-5pt}
    \renewcommand\arraystretch{1.36}
    \begin{tabular}{@{}lcccc@{}}
      \toprule
      Setting      & FID $\downarrow$ & FVD $\downarrow$ 
                  & AbsRel $\downarrow$ & $\delta_1$ $\uparrow$ \\
      \midrule
      Direct add   & 223.2 & 2474.6 & 53.177 & 49.9 \\
      Random-conv &  51.2 & 615.1  & 26.262 & 61.7 \\
      Zero-conv   & \textbf{11.8} & \textbf{99.9} 
                  & \textbf{8.936} & \textbf{91.4} \\
      \bottomrule
      \label{tab:feedback_type}
    \end{tabular}
  \end{subtable}
    \vspace{-20pt}
\end{table*}

\textbf{On future geometric perception,} \ourmethod~surpasses the state-of-the-art depth estimator Marigold~\cite{ke2024repurposing}, achieving the lowest AbsRel (8.936) and the highest threshold accuracies. While Marigold is specialized for high-quality static depth estimation, it operates frame-by-frame and lacks temporal foresight. Consequently, its performance degrades drastically at longer prediction horizons (e.g., AbsRel 39.0 at the 12th frame). In contrast, \ourmethod~benefits from dynamics-informed estimation: by leveraging the video generator's temporal priors, our model anticipates how scene geometry evolves over time. The unified latent space effectively couples pixel dynamics with geometric structural changes, resulting in 4D predictions that are not only sharp but also temporally stable.

\subsection{Zero-Shot Generalization}
To evaluate the robustness of our 4D representations, we test \ourmethod~on the Waymo dataset without fine-tuning. As shown in Tab.~\ref{tab:zero_shot}, our method outperforms Vista~\cite{gao2024vista} in future generation with a 7.5-point lower FID (16.3 vs. 23.8) and improved FVD, demonstrating stronger visual coherence in unseen domains. Crucially, while Vista lacks perception capabilities, \ourmethod~provides accurate zero-shot depth estimation (AbsRel = 9.517), benefiting from the entangled learning of texture and geometry. These results confirm that our unified approach captures fundamental 4D world dynamics that generalize across different driving environments.

\subsection{Ablation Studies}

This section dissects the key components of our 4D world model, validating the design choices for unified learning and spatio-temporal interaction.

\textbf{Impact of Optimization Paradigms (Unified vs. Separate).} We first study how the coupling of modalities affects 4D modeling. As shown in Tab.~\ref{tab:optimization}: \textbf{1)} Image-only training generates reasonable textures but lacks geometric understanding. \textbf{2)} Depth-only training fails to capture complex scene dynamics. \textbf{3)} Separate optimization (Detached-grad) improves depth but degrades generation quality (higher FVD) because geometric constraints are not propagated to the visual stream. \textbf{4)} Our Joint training strategy yields the best performance across all metrics. This confirms that 4D modeling is not a zero-sum game; instead, depth modeling stabilizes visual structure, while visual context refines geometric details, creating a mutually beneficial loop.

\begin{figure*}[t]
	\begin{center}
		\includegraphics[width=0.9\linewidth]{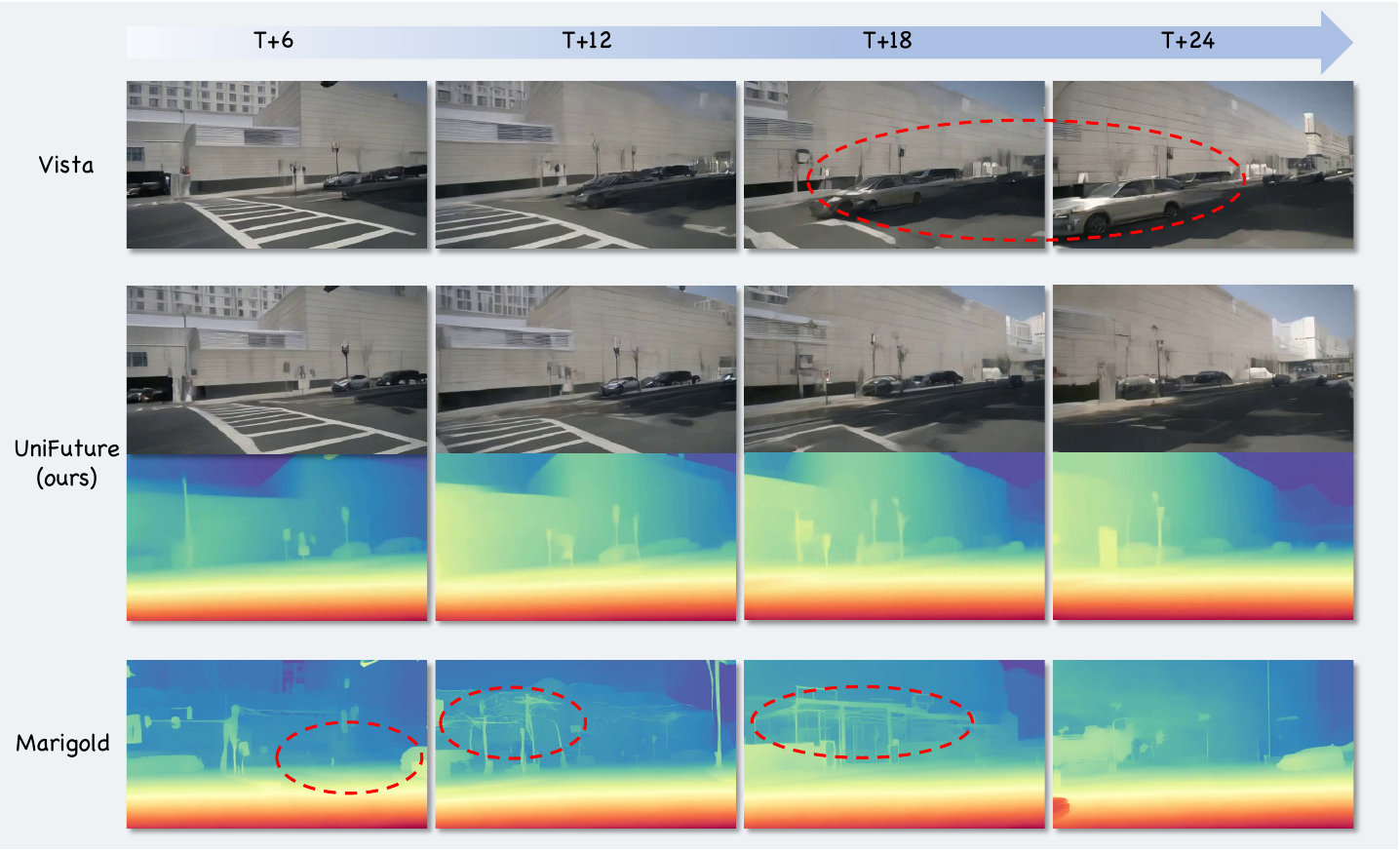}
	\end{center}
        \vspace{-10pt}
	\caption{Qualitative comparisons. Existing DWMs (Vista~\cite{gao2024vista}) generate plausible videos but lack geometric grounding. Specialized depth estimators (Marigold~\cite{ke2024repurposing}) fail to forecast future geometry. In contrast, \ourmethod~delivers coherent 4D predictions, maintaining structural integrity over time.}
	\label{fig:demo}
    \vspace{-5pt}
\end{figure*}

\begin{figure}[t]
	\begin{center}
		\includegraphics[width=0.96\linewidth]{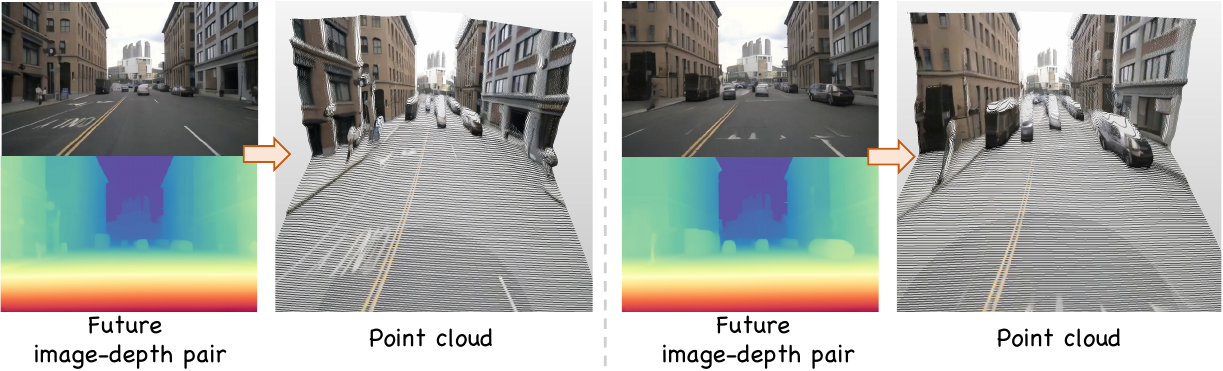}
	\end{center}
        \vspace{-7pt}
	\caption{Visualization of the 4D Point Clouds reconstructed from our predicted future image-depth sequences. This demonstrates \ourmethod's capability to simulate not just 2D videos, but the dynamic evolution of the 3D world.}
	\label{fig:4D_world_model}
      \vspace{-15pt}
\end{figure}

\textbf{Effectiveness of the DLS Scheme (Unified 4D Latent Space).} We evaluate the core idea of mapping image and depth into a shared manifold. As shown in Tab.~\ref{tab:vpd_like}, replacing a conventional convolutional depth decoder with our shared VAE decoder (DLS) significantly improves both generation (FVD: 99.9 vs. 121.1) and perception. This validates that sharing the latent space enforces a tighter coupling between appearance and geometry, which is essential for consistent 4D prediction.

\textbf{Effectiveness of Multi-scale Interactions (MLI).} 
To ensure fine-grained alignment between visual and geometric features, MLI introduces interactions at multiple scales. Tab.~\ref{tab:multiscale} shows that multi-scale feedback significantly outperforms single-scale interaction. This suggests that 4D consistency requires alignment at both high-level semantic scales (for object dynamics) and low-level structural scales (for boundary precision).

\textbf{Role of Bidirectional Feedback.} We analyze the directionality of feature flow in MLI. As shown in Tab.~\ref{tab:feedback}: \textbf{1)} Inside feedback (Geometry-to-Texture) significantly reduces FVD, proving that geometric cues help stabilize video generation. \textbf{2)} Combined with Outside feedback (Texture-to-Geometry), we achieve the best balance. This validates our closed-loop design where geometry constrains appearance and appearance refines geometry.

\textbf{Impact of Feedback Layer Initialization.} Finally, Tab.~\ref{tab:feedback_type} shows that using Zero-conv initialization is critical. Direct addition or random initialization disrupts the pre-trained latent features, leading to collapse. Zero-conv allows the model to gradually learn the mapping between the heterogeneous image and depth spaces, enabling a smooth transition to 4D modeling.

\subsection{Qualitative Analysis: 4D World Reconstruction}

We present qualitative comparisons in Fig.~\ref{fig:demo}, showcasing \ourmethod's superiority in generating realistic and geometrically accurate future scenes. Beyond standard metrics, the ultimate test of a 4D world model is its ability to reconstruct the dynamic 3D world. As illustrated in Fig.~\ref{fig:4D_world_model}, we project our predicted image-depth pairs into 3D space to form 4D Point Clouds. The reconstructed scenes exhibit temporal continuity and structural integrity, with dynamic objects (e.g., moving vehicles) and static backgrounds (e.g., roads, buildings) evolving consistently. This visualization confirms that \ourmethod~successfully transcends 2D video generation, effectively functioning as a robust simulator for the 4D physical world.

\subsection{Controllable Future Scene Evolution}
\begin{figure}[htbp]
	\begin{center}
		\includegraphics[width=0.96\linewidth]{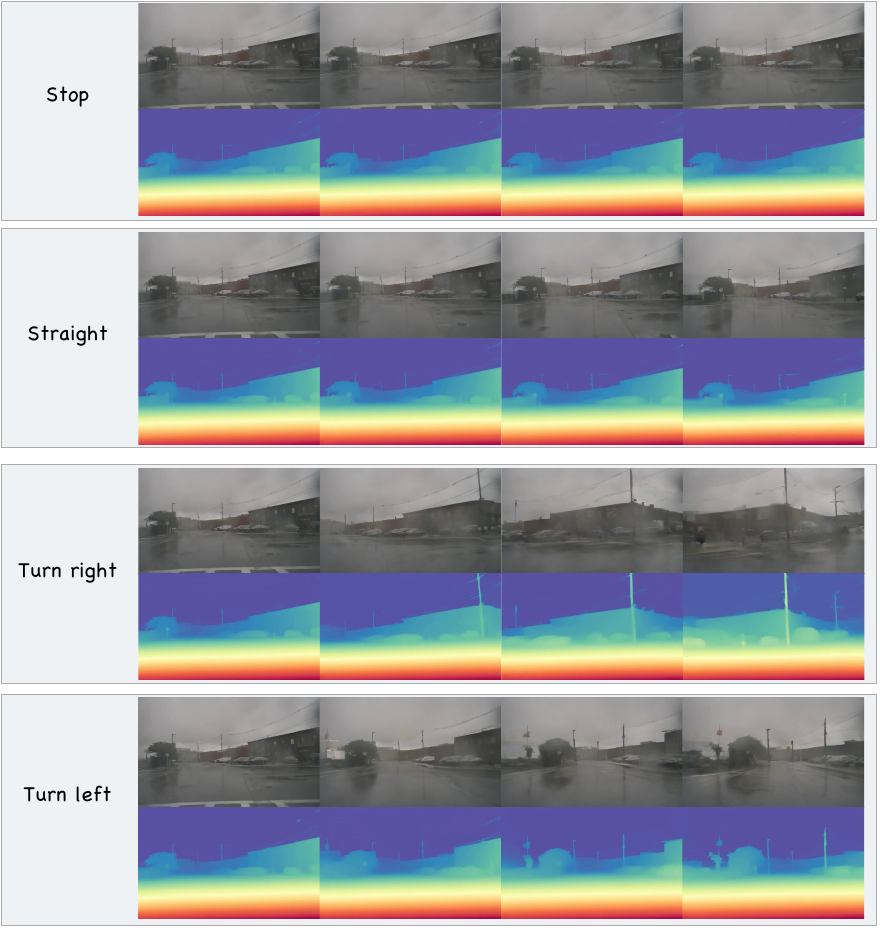}
	\end{center}
        \vspace{-10pt}
	\caption{Controllable future scene evolution with \ourmethod. Given the same starting frame, our method generates diverse future trajectories based on different control commands (e.g., stop, go straight, turn right, turn left) while maintaining high-quality geometric consistency.}
	\label{fig:control}
\end{figure}

A crucial characteristic of world models in autonomous driving is their ability to predict future scene evolution based on control signals. Only with this capability can the world model provide a realistic simulation environment, thereby supporting the development of end-to-end reinforcement learning models. Thus, in this section, we present the results of \ourmethod~for controllable future scene evolution, shown in Fig.~\ref{fig:control}. Starting from the same condition frame, our method can generate different future scenes corresponding to given commands (e.g., go straight, turn right) with high-quality geometry outputs (i.e., depth). These results demonstrate the potential of our method to support downstream tasks, showing the superiority of ~\ourmethod.

\section{CONCLUSIONS}

We propose \ourmethod, a 4D driving world model that integrates future scene generation and depth-aware perception. Through Dual-Latent Sharing (DLS) and Multi-scale Latent Interaction (MLI), our method enables effective cross-modal representation learning and bidirectional refinement, enhancing structural consistency. Experiments show that our method achieves impressive performance, producing high-consistency image-depth pairs with improved realism and geometric alignment. By unifying the future generation and perception, we offer a valuable alternative to world modeling for autonomous driving.

\bibliographystyle{IEEEtran}  
\bibliography{reference}

\end{document}